\def\year{2020}\relax
\documentclass[letterpaper]{article} 
\usepackage{aaai20}  
\usepackage{times}  
\usepackage{helvet} 
\usepackage{courier}  
\usepackage[hyphens]{url}  
\usepackage{graphicx} 
\usepackage{graphicx}  
\usepackage{multirow}
\title{MUSE: Multi-Scale Temporal Features Evolution for Knowledge Tracing}

\author{Chengwei Zhang, Yangzhou Jiang, Wei Zhang, Chengyu Gu
\\ 
Shanghai Jiao Tong University\\ 
\{cwzhang, jiangyangzhou, gcy950912\}@sjtu.edu.cn, 
 mercurialzhang@gmail.com 
}
\begin{document}

\maketitle

\begin{abstract}
Transformer based knowledge tracing model is an extensively studied problem in the field of computer-aided education. By integrating temporal features into the encoder-decoder structure, transformers can processes the exercise information and student response information in a natural way. However, current state-of-the-art transformer-based variants still share two limitations. First, extremely long temporal features cannot well handled as the complexity of self-attention mechanism is $O(n^2)$. Second, existing approaches track the knowledge drifts under fixed a window size, without considering different temporal-ranges. To conquer these problems, we propose MUSE, which is equipped with multi-scale temporal sensor unit,  that takes either local or global temporal features into consideration. The proposed model is capable to capture the dynamic changes in users’ knowledge states at different temporal-ranges, and provides an efficient and powerful way to combine local and global features to make predictions. Our method won the 5-th place over 3,395 teams in the Riiid AIEd Challenge 2020.
\end{abstract}

\section{Introduction}
Recent COVID-19 has forced most countries to temporarily close schools and offline-education is in a tough place. The equity gaps in every country could grow wider since student knowledge over time was hard to trace offline. With the fast evolution of data science, data scientists can help teachers to realize personalized teaching by developing knowledge tracing models and relevant dataset \cite{ednet}.

Existing works mainly focus on diagnosing the knowledge proficiency of students and making prediction of whether they can answer the given exercise correctly. To capture complex relations among exercises and responses over time, the attention mechanism has been widely explored by various knowledge tracing models such as  \cite{sakt,saint,saint+}. However, most of the previous works focused on a fixed time-scale, without paying attention to multi scales. Multi-Scale structures are widely used in computer vision (CV), NLP, and signal processing domains.
It can help the model to capture patterns at different scales and extract robust features. For example, \cite{msmha} introduce the multi-scale structure into self-attention framework and propose a multi-scale multi-head self-attention network. Our work is inspired by their success, and we propose a more powerful network called MUSE to capture the multi-scale time-range features. Our model is composed of two separated modules: MUSE-Local and MUSE-Global, which is designed to capture the  changes of the knowledge in short and long temporal-ranges respectively. The MUSE-Local is a transformer-based variants based on SAINT+ \cite{saint+}. More robust structures like Attentional aggregator \cite{MRIF}, Attentional pooling \cite{din} are introduced to fully capture the short temporal-range dynamic changes under a fixed window size.  The MUSE-Global is a Recurrent Neural Network (RNN) based network. We take full advantage of the characteristic that RNN-structure can be simply extended into unlimited length to capture the extremely long temporal-range global features. These two modules are complementary with each other. Our experiments demonstrate the effectiveness of the ideas and our model has won the 5-th place over 3,395 teams in the Riiid AIEd Challenge 2020.
 
\section{Methodology}

\subsection{Problem Formulation}
Given the history of how a student responded to a set of exercises, MUSE follows SAINT \cite{saint}, and SAINT+ \cite{saint+} that predicts the probability that the student will answer a particular new exercise correctly. In SAINT \cite{saint}, the student activity is recorded as a sequence $I_i,...,I_n$ of interactions $I_i = (E_i,R_i)$, where $E_i$ represents the i-th exercise given to the student with related metadata such as the type of the exercise. Also, the response information $R_i$ denotes the student i-th response to $E_i$ with related metadata such as the duration of time the student took to respond. In this competition, more metadata such as lectures are provided at the same time. We denote them as $L_i$ in this report and the whole interactions can be represented as $X_i = (I_i,L_i)$. The goal of MUSE is to predict the probability $$P(r_k=1 | I_1, I_2,...,E_k)$$ of a student answering the i-th exercise correctly.

\subsection{Feature Engineering}
In this section, we will describe how the input features of our model, as well as the feature engineering of each feature. Here, all the features are simply embedded by either categorical or continuous embeddings proposed by \cite{saint+}. The categorical embeddings are applied by default if not specified.

\subsubsection{Exercise embeddings} The exercise embedding, which contributes to $E_i$,  includes the following features:
\begin{itemize}
	\item{Content and Bundle id}: the content id and bundle id are embedded into a common latent space via a shared embedding matrix.
	\item{Part, Tag, Content answer}: The part, tag and content answer, including different information of an exercise, are separately embedded into an latent vector.
	\item{Position}: The position embeddings of an content in the input sequence. Note that the position embeddings are NOT shared across the exercise sequence and the response sequence in our model.
\end{itemize}

\subsubsection{User embeddings} The user embedding, which contributes to $R_i$, includes the following features:
\begin{itemize}
	\item{Task container id}: the task container id is embedded using categorical embedding.
	\item{Response}: the user’s historical response sequences of possible 0/1 value are embedded into an latent vector.
	\item{Prior questions elapsed time}: elapsed time is an amount of time that a student spent on solving a given exercise. This feature is directly embedded using continuous embedding.
	\item{Lag time}: lag time is defined as the time gap between the end of a previous response and the beginning of current exercise in SAINT+ \cite{saint+}. Here, we define the lag time  as the time gap between beginning of two exercises. We set the maximum lag time as 300 seconds and any time more than that is cut off to 300 seconds. Continuous embedding is used for this feature.
	\item{Prior questions had explanation}: 0/1 value, using categorical embedding.
	\item{Prior questions had been attempted}: this value stands for whether the user has attempted the given exercise. To improve diversity, in MUSE-Local, 0/1 value is applied to represent the state. In MUSE-Global, $1-exp(-x)$ is applied, where x is the cumulative attemption count of the given exercise by the user.
	\item{Position}: as described in the \textbf{Exercise embedding}.
\end{itemize}

\subsubsection{Lecture embeddings} The lecture embedding, which contributes to $L_i$, includes the following features:
\begin{itemize}
\item{Last part}: means whether each part of the lectures has been recorded, which has 7 values.
\item{Last type}: means whether each type of the lectures has been recorded, which has 4 values.
\item{Last tag}: means whether each tag of the lectures has been recorded. Since the lectures share 188 tags, we only select the most frequently used tags (top 14) as the embedding space and denote all other tags as one specific value. Thus the last tag contains 15 values.
\end{itemize}
Note that all the embedded features are directly concatenated together without LayerNorm and Dropout, then linearly transformed into a predefined  common space  (dimension of d\_model). 

\subsubsection{Global features} We also add several global features without any embedding layer, shown as follows:

\begin{itemize}
	\item{Exercise's hotness}: the total count of each content id, cut off by 22000.
	\item{Exercise's hardness}: the accuracy of each content id .
	\item{Exercise's part hardness}: the accuracy of each part.
	\item{User's cumulative response ratio}:  the value is to keep track of the user's preference for each answer. For example, many users will choose C if they do not know the answers.
	\item{User's cumulative correct rate}: the value is to keep track of the accuracy that users have answered the exercise.
	\item{User's cumulative lecture-watching counts}: the value is to keep track of the counts that users have watched lectures.
\end{itemize}

\subsection{Model Architecture}
\begin{figure*}[!t] 
	\centering
	\includegraphics[scale=0.9]{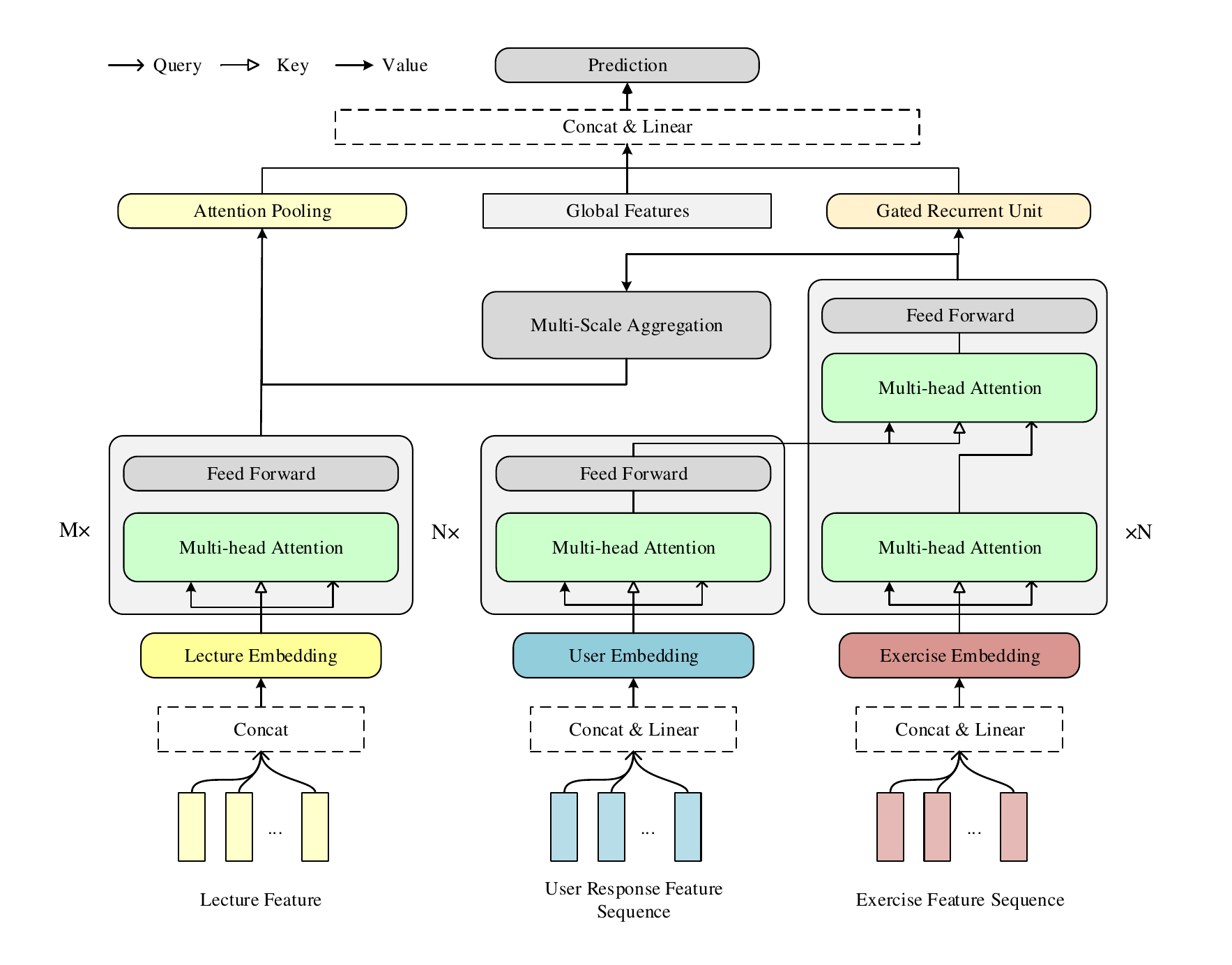} 
	\caption{The architecture of the well-designed MUSE-Local model.}
	\label{figure} 
\end{figure*}

In this section, we will describe our model architecture in details. The overall MUSE-Local is shown in Figure \ref{figure}.

\subsubsection{Self-Attentive Encoder-Decoder}
Following the SAINT+ \cite{saint+}, we directly introduce Self-Attentive Encoder-Decoder to encode the user embeddings and exercise embeddings. The differences are summarized as follows: 
\begin{itemize}
\item[I] We use more sequence features and combine them by concatenating and linear transforming instead of directly adding together. 

\item[II] We add another self-attentive encoder Layer to encode the lecture embeddings separately.

\item[III] The position embeddings of the above encoders and decoders are all not shared.

\item[IV] We have also tried GRU layer \cite{gru} to further model user historical interactions with similar usage in \cite{dien}.  

\end{itemize}

\subsubsection{Multi-Scale Aggregation Layer}
To capture the dynamic changes in user interactions, we introduce the attentional aggregator proposed by \cite{MRIF}. The function of multi-scale interaction aggregation layer is to trace users at different temporal-ranges and form a group of multi-scale user interactions. The attentional aggregator can be computed as follows:
$$Agg([I_{i-w}, I_{i-w+1},...,I_{i+w}]) = \sum_{j=i-w}^{i+w}\alpha_jI_j $$, where w is the window size of the aggregator and $\alpha_j$ is the attention parameter which can be learned during training. In this competition, we use two aggregators as our default settings and set the $w=3$, stride=1 to keep the sequence length unchanged. We did not observe further improvements by enlarging the number of layers and size of $w$. 

\subsubsection{Attention Pooling Layer}
When making the final decisions, we only make predictions for the current step. Therefore we need to transform a 2D feature representations into 1D. Global Average Pooling (GAP) or Global Max Pooling (GMP) can solve the problems, but the result is not satisfactory because the most query-related interaction will be overwhelmed by historical interactions. Here, we use the attention pooling layer proposed by \cite{din}, formulated as follows:
$$\zeta_{Sequence}(Query) =  \sum_{j=1}^{l} a(S_j, Query) $$
where $\{S_1,S_2,...,S_l\}$ is the list of embedding vectors in Sequence, Query is the embedding vector of content id, $l$ is the sequence length. In this way, $\zeta_{Sequence}(Query)$ will vary over different content id. $a(\cdot)$ is a feed-forward network with output as the activate weight. 

At last, to make the final prediction, we concatenate all the pooled features, global features and the GRU output together, which will be transformed into decision space by 3 fully connected layers. 

\begin{table*}[!htb]
	\centering
	\begin{tabular}{lllccl}
		\hline
		& Model       & Number of layers & d\_model & Dropout & Max window size \\ \hline
		1 & MUSE-Local & \multicolumn{1}{l}{3 user encoders + 3 exec. decoders + 2 lect. encoder} & 128 & 0.0 & 200 \\
		2 & MUSE-Local & 3 user encoders + 3 exec. decoders + 2 lect. encoder                      & 128 & 0.0 & 200 \\
		3 & MUSE-Global & 2 GRUs     & 256      & 0.1     & unlimited.       \\
		4 & MUSE-Global & 2 GRUs     & 256      & 0.1     & unlimited.       \\ \hline
	\end{tabular}
	\caption{Models used for final submission. Here exec. decoders means exercise decoders, lect. encoder means lecture encoders.}
	\label{param}
\end{table*}

\begin{table*}[!htb]
	\centering
	\begin{tabular}{l|lll}
		\hline
		\multicolumn{1}{c|}{\multirow{2}{*}{Model Architecture}} & \multicolumn{3}{c}{Train Dataset Size}                                               \\ \cline{2-4} 
		\multicolumn{1}{c|}{}                                    & \multicolumn{1}{c}{10M} & \multicolumn{1}{c}{90M} & \multicolumn{1}{c}{ALL}   \\ \hline
		MUSE-Local w/o Lecture's Branch                          & 0.800                   &                         &                           \\
		MUSE-Local w/o RAM                                       & 0.798                   &                         &                           \\
		MUSE-Local w/o Attention Pooling                         & 0.799                   &                         &                           \\
		MUSE-Local w/o Multi-Scale Aggregation                   & 0.801                   &                         &                           \\
		MUSE-Local w/o Adv. Training                             & 0.801                   &                         &                           \\
		MUSE-Local                                               & 0.802                   & 0.814                   & 0.814                     \\ \hline
		\multicolumn{1}{c|}{MUSE-Global} & \multicolumn{1}{c}{0.801} & \multicolumn{1}{c}{0.813} & \multicolumn{1}{c}{0.813} \\ \hline
		\multicolumn{1}{c|}{MUSE}                                & \multicolumn{1}{c}{}    & \multicolumn{1}{c}{}    & \multicolumn{1}{c}{0.817} \\ \hline
	\end{tabular}
	\caption{Major millstone evaluated by private leaderboard metric ROC AUC, where the final MUSE model are fused by 2 MUSE-Local models and 2 MUSE-Global models. }
	\label{res}
\end{table*}

\subsection{Training Details}

\subsubsection{MUSE-Local} To capture the local sequence features, we use several training techniques as follows:
\begin{itemize}
\item{Normal Training.} We use transformer with d\_model=128 and N=3 and M=2 layers. All the model was trained from scratch. The window size, dropout rate, and batch size are set to 200, 0.0, and 2048 respectively. We use the AdamW \cite{adamw} with lr = 0.001, $\beta_1$ = 0.9, $\beta_2$ = 0.999 and weight decay=0.001. We use the so-called Noam scheme to schedule the learning rate as in \cite{transformer} and set warmup steps to about 8000.
 
\item{Random Answer Masking.}
Inspired by \cite{bert}, MUSE randomly masks some of the user's response from the input sequence, and the objective is to predict the original response of the masked response value based only on its context. The training technique is termed as RAM in this report and the RAM ratio is 25\% by default. 

\item{Adversarial Training.}
Adversarial training, which was originally proposed to defend adversarial examples and enhance the security of machine learning systems \cite{fgsm}, has recently been proved to be effective for improving the generalization of language models \cite{freelb}. Here, we also introduce the adversarial training to improve the generalization of knowledge tracing models. However, since the huge computation cost of adversarial training (it takes 3-30 times longer to form a robust network \cite{advfree}) and the limitation of competition deadline , we only train the MUSE-Local for another 10k steps using techniques introduced by \cite{freelb}, and achieves only (less than)0.001 AUC improvements. We believe that the performance will be better if the models are trained even longer.

\end{itemize}

\subsubsection{MUSE-Global} 
To capture the global sequence features, here we use a 2-layer single-directional GRU \cite{gru} with a dimension of 256 and unlimited window size. The model was trained from scratch using Adam \cite{adam} with lr=0.001, $\beta_1$ = 0.9, $\beta_2$ = 0.999 and weight decay=0.

\subsubsection{MUSE-Fusion} 
To better fuse the local and global models, we train the MUSE-Local and MUSE-Global using 90M of the training data, and simply blend the results on the remaining training data via 5-fold cross validation by LightGBM \cite{lightgbm} with default parameters and early-stopping=100 without any further post-processing.

\section{Experimental Results}
Table \ref{param} is out final submission architectures and Table \ref{res} is our major milestone for the changes in method and framework. 
According to our validation result, the best performing model of MUSE-Local has 3 layers and a latent space dimension of 128. It shows an AUC of 0.814 on the private leaderboard. Since the 13G memory limitations on Kaggle, we only use a dimension of 256 MUSE-Global models and achieve an AUC of 0.813. Dimension of 512 may perform better but consumes much more memory.

\subsubsection{Effect of Lecture's Branch.} The result of MUSE-Local w/o Lecture's Branch shows a decrease of 0.002 AUC when compared with the full MUSE-Local model, proving the effectiveness of the lecture's branch. The result also tells us that watching lectures do have functions to improve the users' ability.

\subsubsection{Effect of Random Answering Masking.}  The result of MUSE-Local w/o RAM shows a significant decrease of 0.004 AUC, proving that the RAM plays an important role in our MUSE model and the context information is crucial.

\subsubsection{Effect of Attention Pooling and Multi-Scale Aggregation Layer.} The result of MUSE-Local w/o Attention Pooling and MUSE-Local w/o Multi-Scale Aggregation shows both part contribute about 0.001 improvements to the final scores.

\subsubsection{Effect of Adversarial Training.} We have already discussed the function of Adversarial Training in Training Details. Here we firmly believe that the performance can be better if the models are trained longer.

\subsubsection{Effect of Multi-Scale Model Fusion.} Our final models are all trained on 90M datasets and we do not found further improvements after adding more training data. Therefore, we use the last nearly 10M dataset to blend all the MUSE-Local and Global models and achieve about 0.003 private LB improvements, which demonstrate the effective of multi-scale model fusion. We can also infer that the transformer-based variants are limited by window size and have drawbacks in dealing with extremely long dependencies.

\section{Conclusion}
In this report, we propose a powerful modified transformer-based model called MUSE for knowledge tracing by automatically aggregating multi-scale temporal features. Experiments show that our solution can perform better than single Transformer method. With this method, we won the 5-th place in the Riiid AIEd Challenge 2020.

\bibliography{bibliography}

\begin{thebibliography}{}

\bibitem[\protect\citeauthoryear{Choi \bgroup et al\mbox.\egroup
  }{2020a}]{saint}
Choi, Y.; Lee, Y.; Cho, J.; Baek, J.; Kim, B.; Cha, Y.; Shin, D.; Bae, C.; and
  Heo, J.
\newblock 2020a.
\newblock Towards an appropriate query, key, and value computation for
  knowledge tracing.
\newblock {\em CoRR} abs/2002.07033.

\bibitem[\protect\citeauthoryear{Choi \bgroup et al\mbox.\egroup
  }{2020b}]{ednet}
Choi, Y.; Lee, Y.; Shin, D.; Cho, J.; Park, S.; Lee, S.; Baek, J.; Bae, C.;
  Kim, B.; and Heo, J.
\newblock 2020b.
\newblock Ednet: A large-scale hierarchical dataset in education.
\newblock In {\em International Conference on Artificial Intelligence in
  Education},  69--73.
\newblock Springer.

\bibitem[\protect\citeauthoryear{Chung \bgroup et al\mbox.\egroup }{2014}]{gru}
Chung, J.; G{\"{u}}l{\c{c}}ehre, {\c{C}}.; Cho, K.; and Bengio, Y.
\newblock 2014.
\newblock Empirical evaluation of gated recurrent neural networks on sequence
  modeling.
\newblock {\em CoRR} abs/1412.3555.

\bibitem[\protect\citeauthoryear{Devlin \bgroup et al\mbox.\egroup
  }{2019}]{bert}
Devlin, J.; Chang, M.; Lee, K.; and Toutanova, K.
\newblock 2019.
\newblock {BERT:} pre-training of deep bidirectional transformers for language
  understanding.
\newblock In {\em NAACL},  4171--4186.

\bibitem[\protect\citeauthoryear{Goodfellow, Shlens, and Szegedy}{2015}]{fgsm}
Goodfellow, I.~J.; Shlens, J.; and Szegedy, C.
\newblock 2015.
\newblock Explaining and harnessing adversarial examples.
\newblock In {\em ICLR}.

\bibitem[\protect\citeauthoryear{Guo \bgroup et al\mbox.\egroup }{2020}]{msmha}
Guo, Q.; Qiu, X.; Liu, P.; Xue, X.; and Zhang, Z.
\newblock 2020.
\newblock Multi-scale self-attention for text classification.
\newblock In {\em AAAI},  7847--7854.

\bibitem[\protect\citeauthoryear{Ke \bgroup et al\mbox.\egroup
  }{2017}]{lightgbm}
Ke, G.; Meng, Q.; Finley, T.; Wang, T.; Chen, W.; Ma, W.; Ye, Q.; and Liu, T.
\newblock 2017.
\newblock Lightgbm: {A} highly efficient gradient boosting decision tree.
\newblock In {\em NeurIPS},  3146--3154.

\bibitem[\protect\citeauthoryear{Kingma and Ba}{2015}]{adam}
Kingma, D.~P., and Ba, J.
\newblock 2015.
\newblock Adam: {A} method for stochastic optimization.
\newblock In {\em ICLR}.

\bibitem[\protect\citeauthoryear{Li, Yang, and Zhang}{2020}]{MRIF}
Li, S.; Yang, D.; and Zhang, B.
\newblock 2020.
\newblock {MRIF:} multi-resolution interest fusion for recommendation.
\newblock In {\em SIGIR},  1765--1768.

\bibitem[\protect\citeauthoryear{Loshchilov and Hutter}{2019}]{adamw}
Loshchilov, I., and Hutter, F.
\newblock 2019.
\newblock Decoupled weight decay regularization.
\newblock In {\em ICLR}.

\bibitem[\protect\citeauthoryear{Pandey and Karypis}{2019}]{sakt}
Pandey, S., and Karypis, G.
\newblock 2019.
\newblock A self-attentive model for knowledge tracing.
\newblock {\em CoRR} abs/1907.06837.

\bibitem[\protect\citeauthoryear{Shafahi \bgroup et al\mbox.\egroup
  }{2019}]{advfree}
Shafahi, A.; Najibi, M.; Ghiasi, A.; Xu, Z.; Dickerson, J.~P.; Studer, C.;
  Davis, L.~S.; Taylor, G.; and Goldstein, T.
\newblock 2019.
\newblock Adversarial training for free!
\newblock In {\em NeurIPS},  3353--3364.

\bibitem[\protect\citeauthoryear{Shin \bgroup et al\mbox.\egroup
  }{2020}]{saint+}
Shin, D.; Shim, Y.; Yu, H.; Lee, S.; Kim, B.; and Choi, Y.
\newblock 2020.
\newblock {SAINT+:} integrating temporal features for ednet correctness
  prediction.
\newblock {\em CoRR} abs/2010.12042.

\bibitem[\protect\citeauthoryear{Vaswani \bgroup et al\mbox.\egroup
  }{2017}]{transformer}
Vaswani, A.; Shazeer, N.; Parmar, N.; Uszkoreit, J.; Jones, L.; Gomez, A.~N.;
  Kaiser, L.; and Polosukhin, I.
\newblock 2017.
\newblock Attention is all you need.
\newblock In {\em NeurIPS},  5998--6008.

\bibitem[\protect\citeauthoryear{Zhou \bgroup et al\mbox.\egroup }{2018}]{din}
Zhou, G.; Zhu, X.; Song, C.; Fan, Y.; Zhu, H.; Ma, X.; Yan, Y.; Jin, J.; Li,
  H.; and Gai, K.
\newblock 2018.
\newblock Deep interest network for click-through rate prediction.
\newblock In {\em KDD},  1059--1068.

\bibitem[\protect\citeauthoryear{Zhou \bgroup et al\mbox.\egroup }{2019}]{dien}
Zhou, G.; Mou, N.; Fan, Y.; Pi, Q.; Bian, W.; Zhou, C.; Zhu, X.; and Gai, K.
\newblock 2019.
\newblock Deep interest evolution network for click-through rate prediction.
\newblock In {\em AAAI},  5941--5948.

\bibitem[\protect\citeauthoryear{Zhu \bgroup et al\mbox.\egroup
  }{2020}]{freelb}
Zhu, C.; Cheng, Y.; Gan, Z.; Sun, S.; Goldstein, T.; and Liu, J.
\newblock 2020.
\newblock Freelb: Enhanced adversarial training for natural language
  understanding.
\newblock In {\em ICLR}.

\end{thebibliography}
\end{document}